\documentclass[10pt,twocolumn,letterpaper]{article}

\usepackage{wacv}
\usepackage{times}
\usepackage{epsfig}
\usepackage{graphicx}
\usepackage[utf8]{inputenc}
\usepackage[english]{babel}
\usepackage{amsmath,amsfonts,amsthm}
\usepackage{csquotes}
\usepackage{xcolor}
\newcommand{\commentMR}[1]{#1}

\newcommand{\commentRC}[1]{#1}
\newcommand{\commentTH}[1]{#1}
\newcommand{\changed}[1]{#1}
\newcommand{\commentPS}[1]{#1}

\newcommand{\comment}[1]{#1}

\newcommand{\intr}{{in}}
\newcommand{\bdr}{{bd}}

\newcommand\Tstrut{\rule{0pt}{2.6ex}}         % = `top' strut
\newcommand\Bstrut{\rule[-0.9ex]{0pt}{0pt}}   % = `bottom' strut

\newcommand{\IoU}{\mathit{IoU}}
\newcommand{\mIoU}{\mathit{mIoU}}
\newcommand{\sIoU}{\mathit{IoU}_{\mathrm{adj}}}
\newcommand{\dsc}{\mathit{Dice}}

\DeclareMathOperator*{\argmax}{arg\, max}

\usepackage[pagebackref=true,breaklinks=true,colorlinks,bookmarks=false]{hyperref}
\usepackage[nameinlink,noabbrev]{cleveref}

\wacvfinalcopy % *** Uncomment this line for the final submission

\def\wacvPaperID{879} % *** Enter the wacv Paper ID here

\ifwacvfinal\fi
\begin{document}

%%%%%%%%% TITLE
\title{Prediction Error Meta Classification in Semantic Segmentation: Detection via Aggregated Dispersion Measures of Softmax Probabilities}
\author{Matthias Rottmann$^{1}$, Pascal Colling$^{1}$, Thomas Paul Hack$^{2}$,\\Robin Chan$^{1}$, Fabian H\"uger$^{3}$, Peter Schlicht$^{3}$ and Hanno Gottschalk$^{1}$ \\
\\
$^{1}$University of Wuppertal, School of Mathematics and Natural Sciences\\
$^{2}$University Leipzig, Faculty of Physics\\
$^{3}$Volkswagen Group Innovation, COI Automation, Architecture and AI-Technologies\\
\tt\small{\{\href{mailto:rottmann@uni-wuppertal.de}{rottmann},\href{mailto:pascal.colling@uni-wuppertal.de}{pascal.colling},\href{mailto:robin.chan@uni-wuppertal.de}{robin.chan},\href{mailto:hanno.gottsch@uni-wuppertal.de}{hanno.gottschalk}\}@uni-wuppertal.de}\\
\tt\small{\href{mailto:thomas-paul.hack@itp.uni-leipzig.de}{thomas-paul.hack}@itp.uni-leipzig.de}\\
\tt\small{\{\href{mailto:peter.schlicht@volkswagen.de}{peter.schlicht},\href{mailto:fabian.hueger@volkswagen.de}{fabian.hueger}\}@volkswagen.de} 
}

% \title{\LaTeX\ Author Guidelines for WACV Proceedings}

% Authors at the same institution
%\author{First Author \hspace{2cm} Second Author \\
% Authors at different institutions
% \author{First Author \\
% \and
% Second Author \\

\maketitle
\ifwacvfinal\thispagestyle{empty}\fi

%%%%%%%%% ABSTRACT
\begin{abstract}
We present a method that ``meta'' classifies whether segments predicted by a semantic segmentation neural network intersect with the ground truth. \comment{For this purpose}, we employ measures of dispersion for predicted pixel-wise class probability distributions, like classification entropy, that yield heat maps of the input scene's size. We aggregate these dispersion measures segment-wise and derive metrics that are well-correlated with the segment-wise $\IoU$ of prediction and ground truth. \changed{This procedure yields an almost plug and play post-processing tool to rate the prediction quality of semantic segmentation networks on segment level.} 
This is especially relevant \changed{for monitoring neural networks} in online applications like automated driving or medical imaging where reliability is of utmost importance. 
\changed{In our tests, we use publicly available state-of-the-art networks trained on the Cityscapes dataset and the BraTS2017 data set and analyze the predictive power of different metrics \comment{as well as} different sets of metrics.} To this end, we compute logistic LASSO regression fits for the task of classifying $\IoU=0$ vs.\ $\IoU > 0$ per segment and obtain \changed{AUROC values of up to $91.55\%$}. We complement these tests with linear regression fits to predict the segment-wise $\IoU$ and obtain prediction standard deviations of down to $0.130$ as well as $R^2$ values of up to \changed{$84.15\%$}. We show that these results clearly outperform standard
approaches.
\end{abstract}

%%%%%%%%% BODY TEXT
\section{Introduction}

In recent years, deep learning has outperformed other classes of predictive models in many applications. In some of these, \eg autonomous driving or diagnostics in medicine, the reliability of a prediction is of highest interest. In classification tasks, the thresholding on the highest softmax probability or thresholding on the entropy of the classification distributions (softmax output) are commonly used approaches to detect false predictions of neural networks, see \eg~\cite{DBLP:journals/corr/HendrycksG16c,DBLP:journals/corr/LiangLS17}. Metrics like classification entropy or the highest softmax probability are usually combined with model uncertainty (Monte-Carlo (MC) dropout inference) and sometimes input uncertainty, cf.~\cite{Gal:2016:DBA:3045390.3045502} and \cite{DBLP:journals/corr/LiangLS17}, respectively. These approaches have proven to be practically efficient for detecting uncertainty. Such methods have also been transferred to semantic segmentation tasks. See also \cite{oberdiek2018} for further uncertainty metrics. The work presented in \cite{DBLP:journals/corr/KendallBC15} makes use of MC dropout to model the uncertainty of segmentation networks and also shows performance improvements in terms of segmentation accuracy. This approach was applied in other works to model the uncertainty and filter out predictions with low reliability, cf.~\eg~\cite{Kampffmeyer2016SemanticSO,DBLP:journals/corr/abs-1807-10584}. In \cite{huang2018efficient} this line of research was further developed to detect spacial and temporal uncertainty in the semantic segmentation of videos.

In this work we establish an approach for efficiently meta-classifying whether an inferred segment (representing a predicted object) of a semantic segmentation
%\commentPS{(i.e., a connected component segmented into the same semantic class - in particular inferred semantic object masks)}
intersects with the ground truth or not, \commentPS{as similarly proposed for classification problems in \cite{DBLP:journals/corr/HendrycksG16c}}. The term \commentPS{\textit{meta classification} has been used in the context of classical machine learning} for learning the weights for each member of a committee of classifiers \cite{Lin2003}. In terms of deep learning we use \commentPS{it} as a shorthand to distinguish between a network's own classification and the classification whether a prediction is \commentPS{``true'' or ``false''}. In contrast to \commentPS{the work cited above, we aim at judging} the statistical reliability of each segment \commentPS{inferred} by the neural network. To this end, dispersion measures, like \commentPS{entropy, are applied to the softmax probabilities (the networks output) on pixel level yielding dispersion heat maps}. We aggregate \commentPS{these heat maps over predicted segments alongside with other quantities derived from the network's prediction like the segment's size and predicted class. From this, we  construct per-segment metrics}. A commonly used performance measure for the quality of a segmentation is the intersection over union ($\IoU$ a.k.a.\ Jaccard index \cite{Jaccard12similarityCoefficient}) of prediction and ground truth. We use the constructed metrics as inputs to logistic regression models for meta classifying, \commentPS{whether an inferred segment's $\IoU$ vanishes or not, i.e., predicting $\IoU=0$ or $\IoU>0$. Also, we use linear regression models for predicting a segment's $\IoU$ directly, thus obtaining statements} about the reliability of the network's prediction. \commentMR{Our method only uses the softmax output of a semantic segmentation network and the corresponding ground truth. It can be equipped with any heat map obtained from pixel-wise uncertainty measures. Thus, any work on uncertainty quantification for semantic segmentation that yields new improved dispersion heat maps can be seamlessly integrated and leverages our method. Hence, we also provide a framework to evaluate the quality of pixel-wise uncertainty measures for semantic segmentation.}

\changed{In our tests we use two publicly available datasets: Cityscapes \cite{cityscapes} for the semantic segmentation of street scenes and BraTS2017 \cite{Bakas2017, 6975210} for brain tumor segmentation. For each of the two datasets we employ two state-of-the-art networks. We perform tests on validation sets and demonstrate that our segment-wise metrics are well correlated with the $\IoU$; thus they are suitable for detecting false positives \commentPS{on segment level}. For logistic regression fits we obtain \commentPS{values of up to $91.55\%$ for the area under curve corresponding to the receiver operator characteristic curve (AUROC, see  \cite{DBLP:conf/icml/DavisG06})}. Predicting the segment-wise $\IoU$ via linear regression we obtain prediction standard deviations of down to $0.130$ and $R^2$ values of up to $84.15\%$.}

\section{Pixel-wise dispersion metrics and aggregation over segments} \label{sec:pixel}

% \begin{figure}
% \begin{floatrow}
% \ffigbox{%
%   \centerline{\includegraphics[width=.249\textwidth]{./figs/entropy.png}\includegraphics[width=.249\textwidth]{./figs/difference.png}}
 
% }{%
%   \caption{Visualization of entropy (cf.~\eqref{eqa:entropy}, left panel) and difference (cf.~\eqref{eqa:alt}, right hand panel) for three variables $p_1$, $p_2$ and $p_3$ treating $p_3$ implicitly via $p_1+p_2+p_3 = 1$. \label{fig:dispersion}}%
% }
% \ffigbox{%
%   \centerline{\includegraphics[width=.249\textwidth]{./figs/wtal_xc.jpg}\includegraphics[width=.249\textwidth]{./figs/wtal_mn.jpg}}
% }{%
%   \caption{Example of segmentation (top line) and heat map $D_z$ (bottom line) for Xception65 (left column) and MobilenetV2 (right hand column). Original image is not part of the Cityscapes dataset. \label{fig:heat}}%
% }
% \end{floatrow}
% \end{figure}

% \begin{figure}
%     \centering
%     \includegraphics[width=.42\linewidth]{./figs/entropy_crop.png}\hspace{1ex}\includegraphics[width=.42\linewidth]{./figs/difference_crop.png}
%     \caption{Visualization of entropy (cf.~\eqref{eqa:entropy}, left panel) and difference (cf.~\eqref{eqa:alt}, right hand panel) for three variables $p_1$, $p_2$ and $p_3$ treating $p_3$ implicitly via $p_1+p_2+p_3 = 1$.}
%     \label{fig:dispersion}
% \end{figure}

\begin{figure}
    \begin{center}
    \includegraphics[width=.49\linewidth]{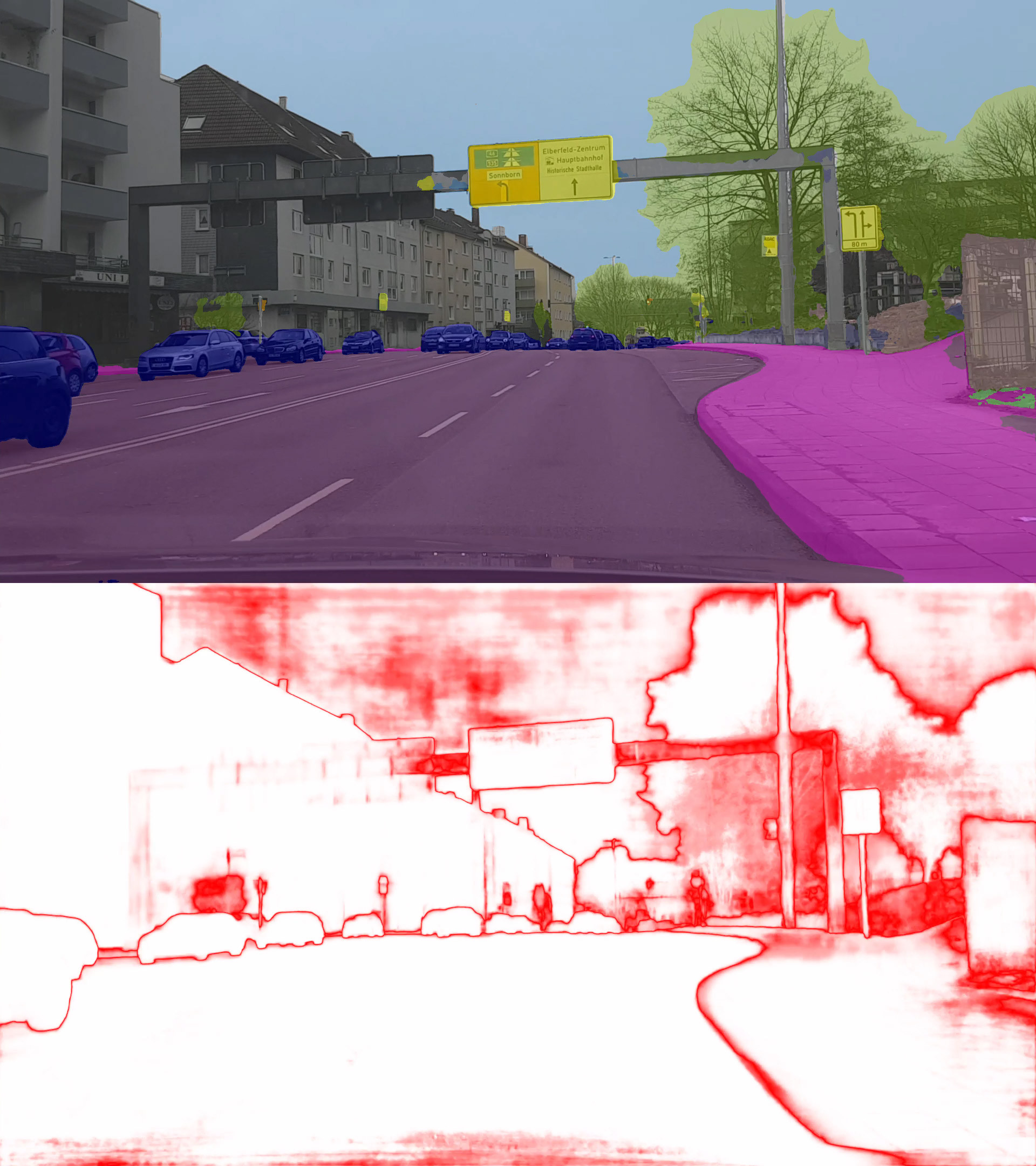}\includegraphics[width=.49\linewidth]{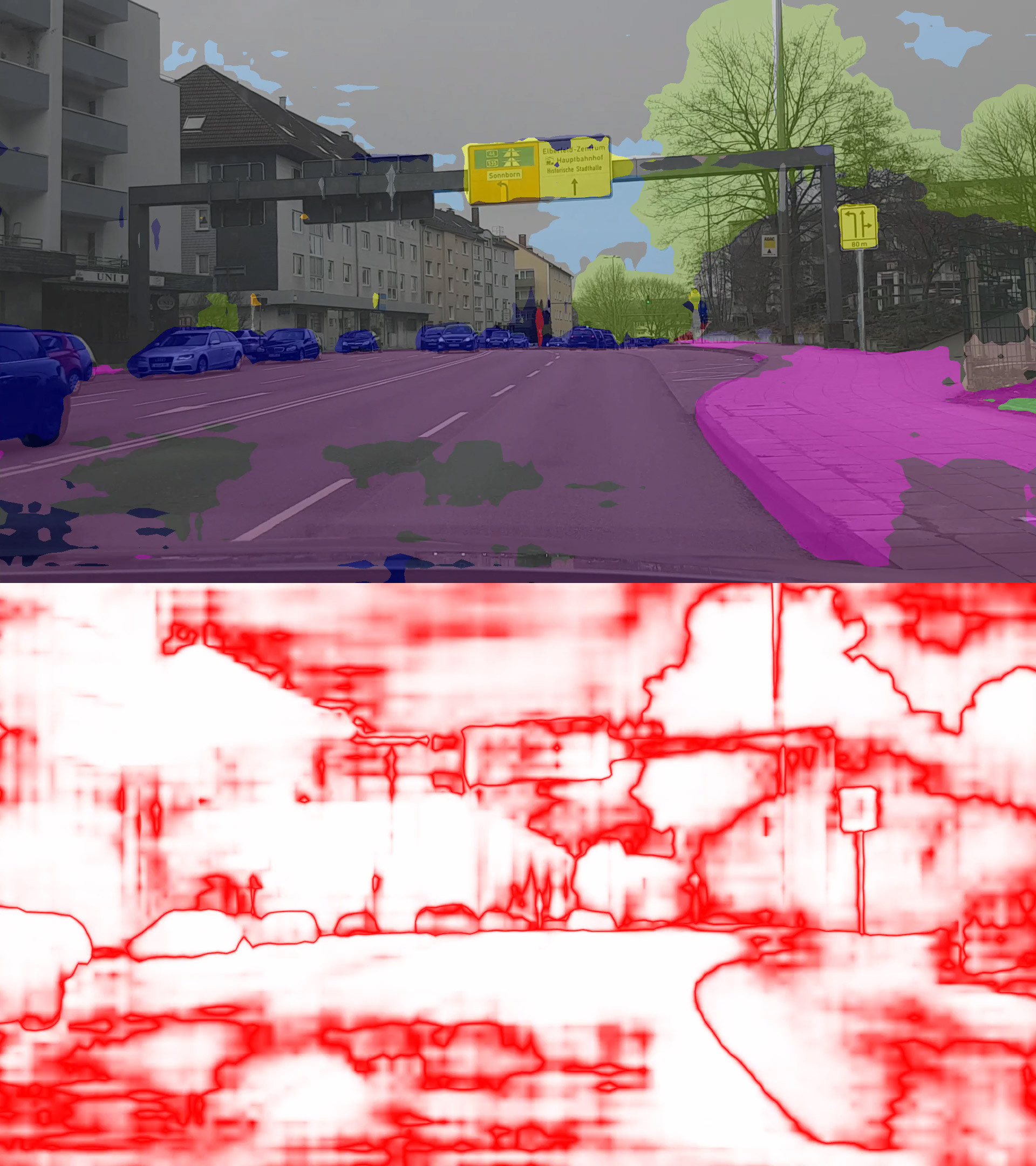}
    \end{center}
    \caption{\comment{Segmentation example} % Example of segmentation
    (top line) and heat map $D_z$ (bottom line) for Xception65 (left column) and MobilenetV2 (right column). Original image is not part of the Cityscapes data set.}
    \label{fig:heat}
\end{figure}
A segmentation network with a softmax output layer can be seen as a statistical model that provides for each pixel $z$ of the image a probability distribution $f_z(y|x,w)$ on the $q$ class labels $y\in\mathcal{C}=\{y_1,\ldots,y_q\}$, given the weights $w$ and the data $x$. The predicted class in $z$ is then given by
\begin{equation}
\hat y_z(x,w)=\argmax_{y\in\mathcal{C}}f_z(y|x,w).
\end{equation}
Dispersion or concentration measures quantify the degree of randomness in \commentPS{$f_z(y|x,w)$. Here,} we consider two \commentPS{of those} measures: \commentPS{\textit{entropy} $E_z$ (also known as \textit{Shannon information}}~\cite{shannon1948}) and \commentPS{\textit{difference in probability} $D_z$, i.e. the difference between the two largest softmax values}:
\begin{align}
\label{eqa:entropy}
E_z(x,w)&=-\frac{1}{\log(q)}\sum_{y\in \mathcal{C}}f_z(y|x,w)\log f_z(y|x,w) \, , \\
\label{eqa:alt}
D_z(x,w)&= 1-f_z(\hat y_z(x,w)|x,w)\nonumber\\
&~~~~~~~~~~~~~~+\max_{y\in\mathcal{C}\setminus\{\hat y_z(x,w)\}} f_z(y|x,w) \, . 
\end{align}
For better comparison, both quantities have been written as dispersion measures and been normalized to \commentPS{the interval $[0,1]$: One has $E_z=D_z=1$ for the equiprobability distribution $f_z(y|x,w)=\frac{1}{q}$, $y\in\mathcal{C}$, and $E_z=D_z=0$ on the deterministic probability distribution ($f_z(y|x,w)=1$ for one class and $0$ otherwise)}. For the discussion of further dispersion measures, cf.~\cite{Cowell1995}.
% \Cref{fig:dispersion} displays these quantities for three class probability distributions. 
The \commentPS{most direct} method of uncertainty quantification on an image is the heat mapping of a dispersion measure \commentPS{as in} \cref{fig:heat}.

%\section{Aggregation over Segments}

For a given image $x$ we denote by $\hat{\mathcal K}_x$ the set of connected components (segments) in the predicted segmentation $\hat{\mathcal S}_x=\{\hat y_z(x,w) | z\in x\}$ \commentPS{(omitting the dependence on the weights $w$)}. Analogously we denote by ${\mathcal K}_x$ the set of connected components in the ground truth ${\mathcal S}_x$. For each $k\in \hat{\mathcal K}_x$, we define the following quantities:
\begin{itemize}
    \setlength\itemsep{-0.05em}
    \item the interior $k_\intr\subset k$ where a pixel $z$ is an element of $k_\intr$ if all eight neighbouring pixels are an element of $k$
    \item the boundary $k_\bdr=  k \setminus k_\intr$
    \item the intersection over union $\IoU$: let ${\mathcal K}_x|_k$ be the set of all $k'\in {\mathcal K}_x$ that have non-trivial intersection with $k$ and whose class label equals the predicted class for $k$, then
    $$
    \IoU(k) = \frac{|k \cap K'|}{|k \cup K'|}\,,\qquad K' = \bigcup_{k' \in {\mathcal K}_x|_k} k'
    $$
    \item adjusted \comment{intersection over union} $\sIoU$: let $Q = \{ q \in \hat{\mathcal{K}}_x: q \cap K' \neq \emptyset \} $, for reasons explained in \cref{sec:sIou} we use in our tests
    $$
    \sIoU(k) = \frac{|k \cap K'|}{|k \cup (K' \setminus Q)|}
    $$
    \item the pixel sizes $S=|k|$, $S_\intr=|k_\intr|$, $S_\bdr=|k_\bdr|$
    \item the mean entropies $\bar E$, $\bar E_\intr$, $\bar E_\bdr$ defined as 
    $$\bar E_\sharp(k) = \frac{1}{S_\sharp} \sum_{z\in k_\sharp} E_z(x)\,,\qquad \sharp\in \{\_,in,bd\}$$
    \item the mean distances $\bar D$, $\bar D_\intr$, $\bar D_\bdr$ defined in analogy to the mean entropies
    \item the relative sizes $\tilde S = S/S_\bdr$, $\tilde S_\intr = S_\intr/S_\bdr$
    \item the relative mean entropies $\tilde {\bar E} = \bar E \tilde S$, $\tilde {\bar E}_\intr = \bar E_\intr \tilde S_\intr$, and relative mean distances $\tilde {\bar D} = \bar D \tilde S$, $\tilde {\bar D}_\intr = \bar D_\intr \tilde S_\intr$
\end{itemize}
Typically, \commentPS{$E_z$ and $D_z$ are large for $z\in k_\bdr$}. This motivates the separate treatment of interior and boundary measures. With the exception of $\IoU$ and $\sIoU$, all scalar quantities defined above can be computed without the knowledge of the ground truth. Our aim is to analyze to which extent they are able to predict \commentPS{$\sIoU$}.

\section{Numerical Experiments: Street Scenes} \label{sec:numexp}

% \begin{figure}[t]
% \begin{floatrow}
% \ffigbox{%
%   \centerline{\includegraphics[width=.249\textwidth]{./figs/XC_regression1.png} \hfill \includegraphics[width=.249\textwidth]{./figs/MN_regression1.png}}%
% }{%
%   \caption{$\sIoU$ vs. predicted $\sIoU$ for all connected components predicted by Xception65 (left) and MobilenetV2 (right). Dot sizes are proportional to $S$. \label{fig:regr} }%
% }
% \capbtabbox{%
% \scalebox{0.76}{\begin{tabular}{|l|r|r||l|r|r|}
% \hline
%                       & XC       & MN       &                        & XC       & MN       \\
% \hline
% $\bar E$               & -0.70139 & -0.70162 & $\bar D$               & -0.85211 & -0.84858\Tstrut\\
% $\bar E_\bdr$          & -0.44065 & -0.41845 & $\bar D_\bdr$          & -0.60308 & -0.52163 \\
% $\bar E_\intr$         & -0.71623 & -0.69884 & $\bar D_\intr$         & -0.85458 & -0.82171 \\
% $\tilde{\bar E}$       &  0.31219 &  0.36261 & $\tilde{\bar D}$       &  0.22797 &  0.30245 \\
% $\tilde{\bar E}_\intr$ &  0.39195 &  0.42806 & $\tilde{\bar D}_\intr$ &  0.29279 &  0.35131 \\
% \hline
% $S$                    &  0.30442 & 0.47978  & $\tilde S$             &  0.50758 &  0.71071\Tstrut\\
% $S_\bdr$               &  0.44625 & 0.62713  & $\tilde S_\intr$       &  0.50758 &  0.71071 \\
% $S_\intr$              &  0.30201 & 0.47708\Bstrut &&& \\
% \hline
% \end{tabular}}
% }{\caption{Correlation coefficients $\rho$ with respect to $\sIoU$. Results are computed on the Cityscapes validation set, XC: DeepLabv3+Xception65 and MN: DeepLabv3+MobilenetV2. \label{tab:correlations}}%
% }
% \end{floatrow}
% \end{figure}
\begin{figure}
    \centering
    \includegraphics[width=.49\linewidth]{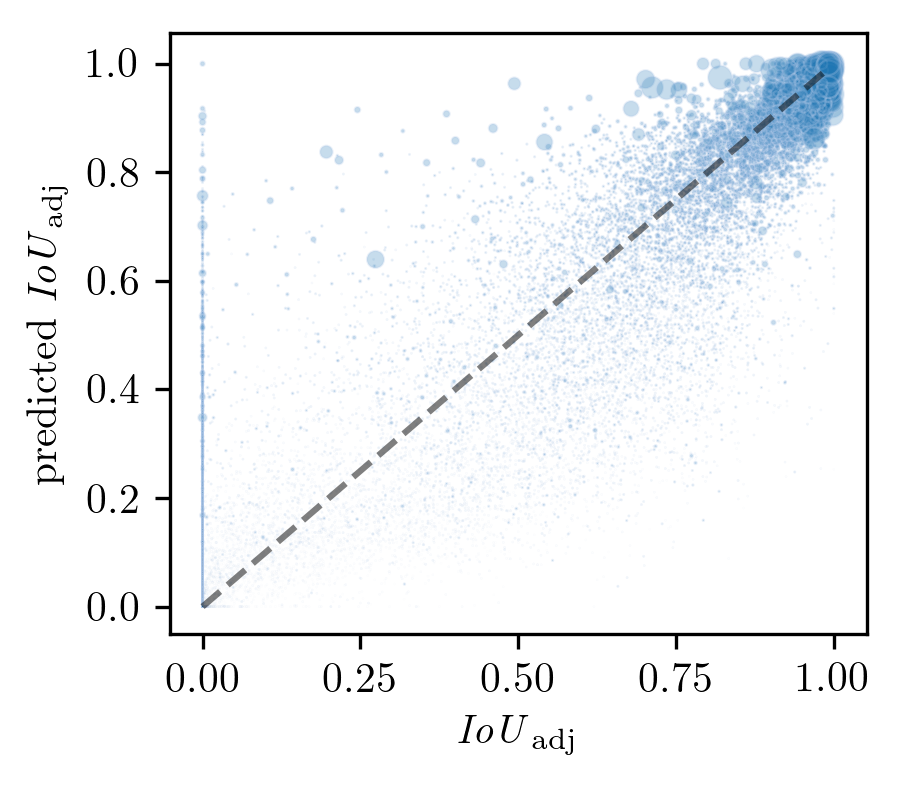} \hfill \includegraphics[width=.49\linewidth]{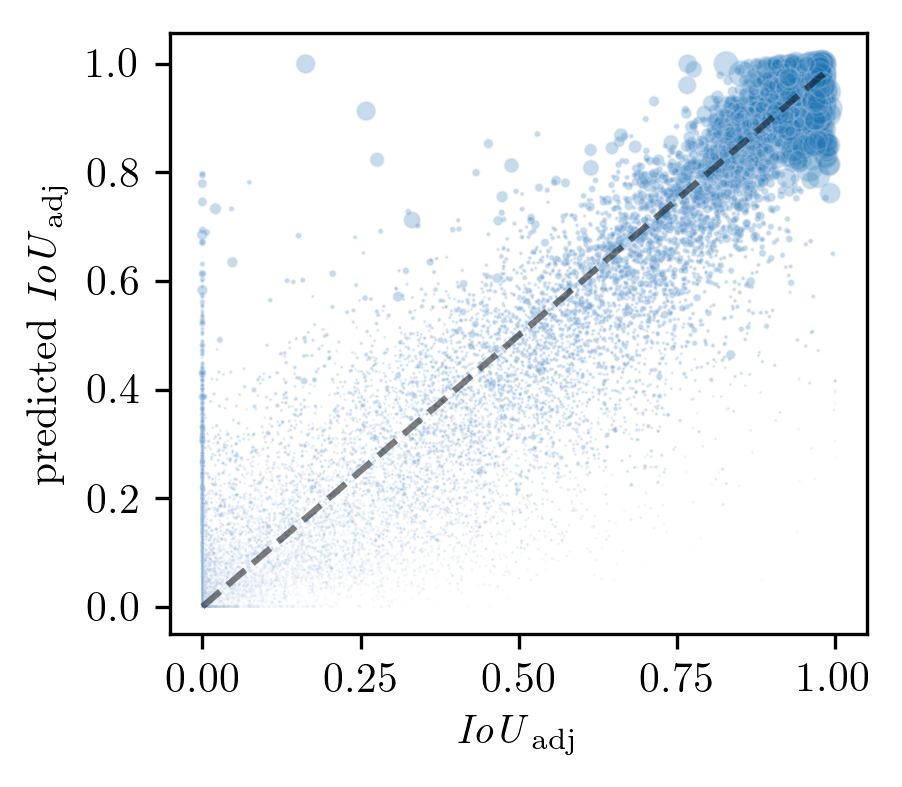}
    \caption{$\sIoU$ vs. predicted $\sIoU$ for all connected components predicted by Xception65 (left) and MobilenetV2 (right). Dot sizes are proportional to $S$.} 
    \label{fig:regr}
\end{figure}
\begin{table}
    \begin{center}
    \scalebox{0.76}{
    \begin{tabular}{|l|r|r||l|r|r|}
    \hline
                          & XC       & MN       &                        & XC       & MN       \\
    \hline
    $\bar E$               & -0.70139 & -0.70162 & $\bar D$               & -0.85211 & -0.84858\Tstrut\\
    $\bar E_\bdr$          & -0.44065 & -0.41845 & $\bar D_\bdr$          & -0.60308 & -0.52163 \\
    $\bar E_\intr$         & -0.71623 & -0.69884 & $\bar D_\intr$         & -0.85458 & -0.82171 \\
    $\tilde{\bar E}$       &  0.31219 &  0.36261 & $\tilde{\bar D}$       &  0.22797 &  0.30245 \\
    $\tilde{\bar E}_\intr$ &  0.39195 &  0.42806 & $\tilde{\bar D}_\intr$ &  0.29279 &  0.35131 \\
    \hline
    $S$                    &  0.30442 & 0.47978  & $\tilde S$             &  0.50758 &  0.71071\Tstrut\\
    $S_\bdr$               &  0.44625 & 0.62713  & $\tilde S_\intr$       &  0.50758 &  0.71071 \\
    $S_\intr$              &  0.30201 & 0.47708\Bstrut &&& \\
    \hline
    \end{tabular}
    }
    \end{center}
    \caption{Correlation coefficients $\rho$ with respect to $\sIoU$. Results are computed on the Cityscapes validation set, XC: DeepLabv3+Xception65 and MN: DeepLabv3+MobilenetV2.}
    \label{tab:correlations}
\end{table}
\begin{figure*}[t]
\centerline{\includegraphics[width=.495\textwidth]{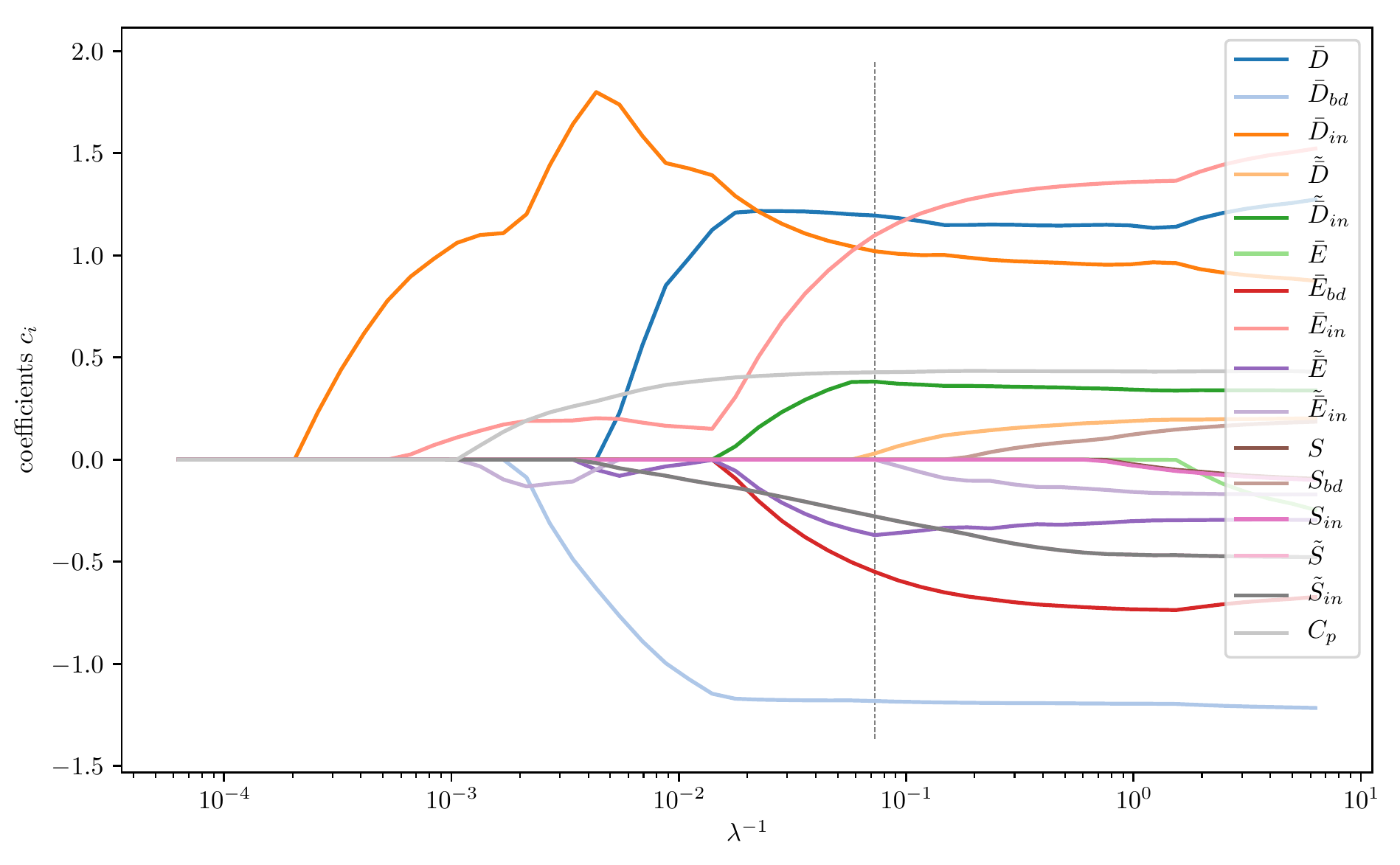} \hfill \includegraphics[width=.495\textwidth]{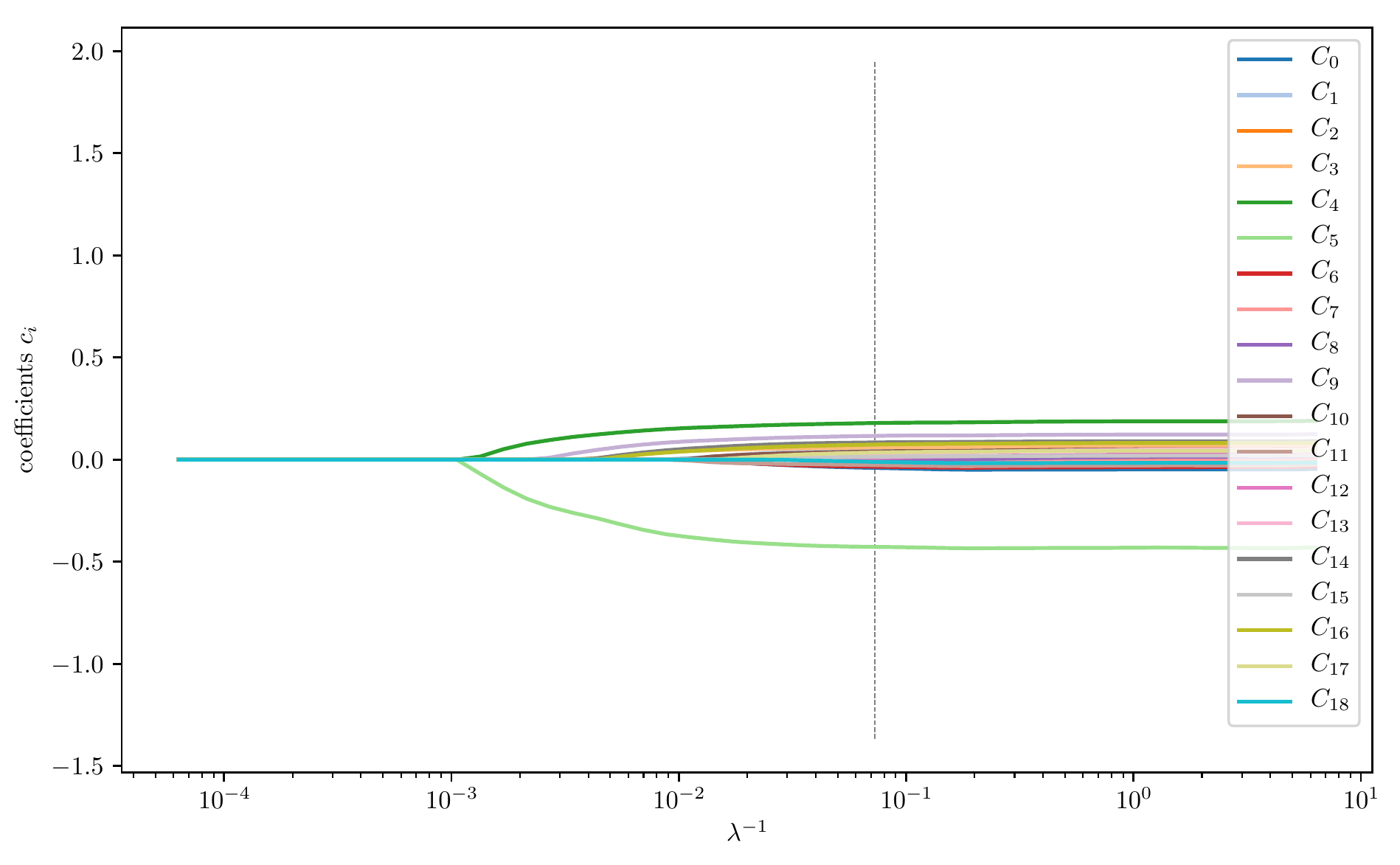}}
\centerline{\includegraphics[width=.495\textwidth]{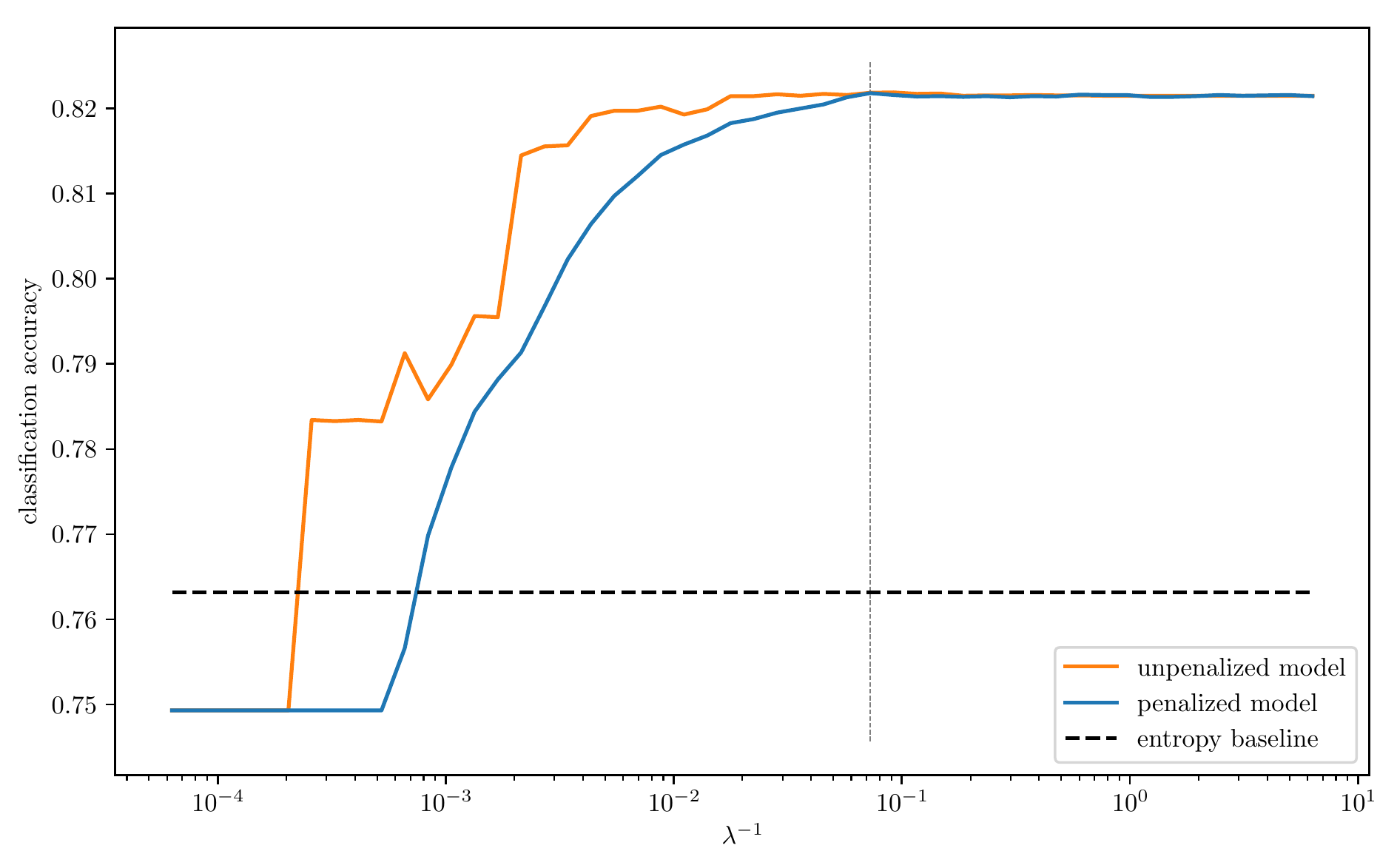} \hfill \includegraphics[width=.495\textwidth]{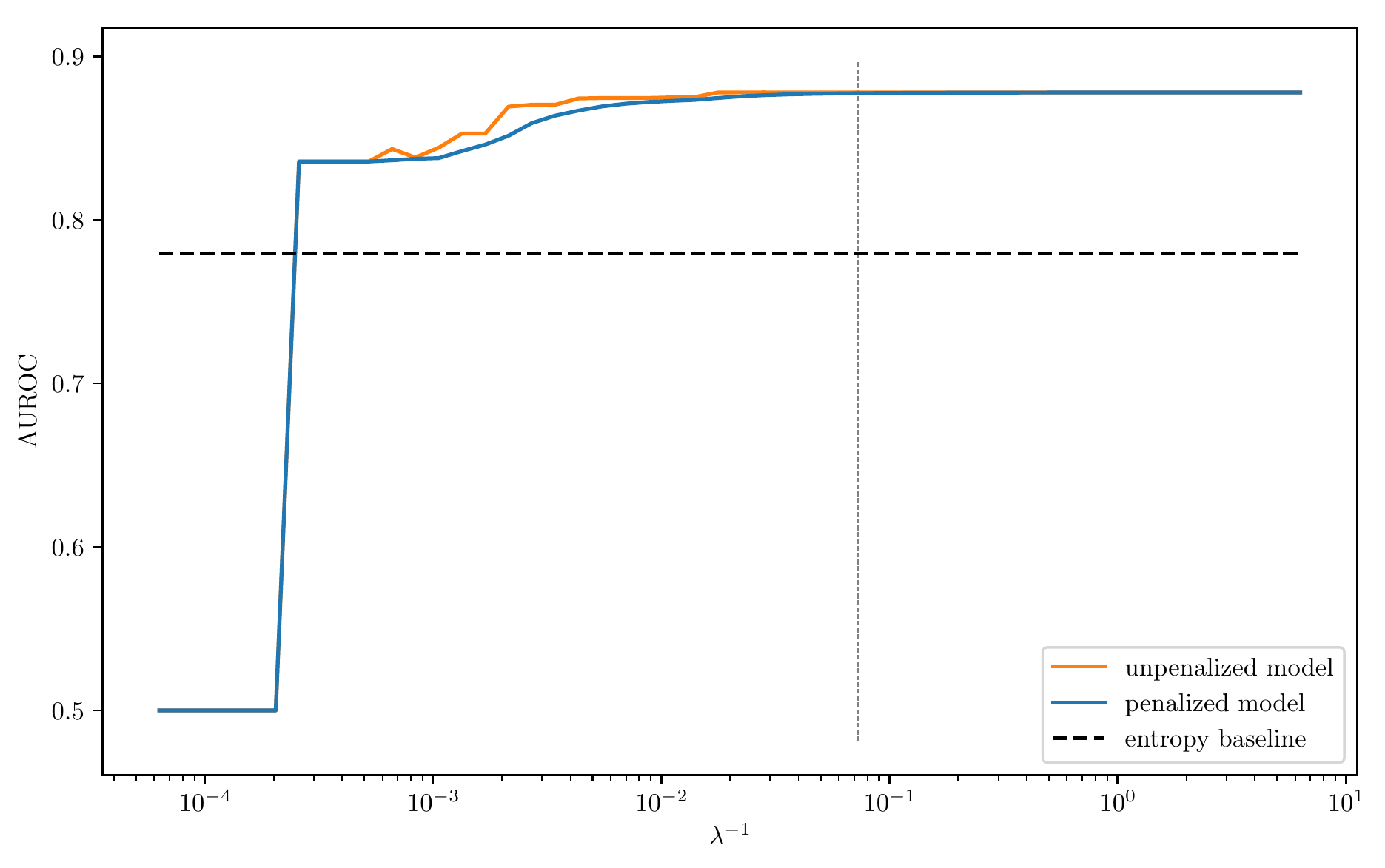}}
\caption{Results for the meta classification task $\sIoU=0,>0$ for predictions obtained from the Xception65 net. (Top left): the weights coefficients for the 15 metrics computed with LASSO fits as function of $\lambda^{-1}$, $C_p$ denotes the maximum of the absolute values of all weight coefficients for predicted classes. (Top right): like top left but showing coefficients for the 18 predicted classes. (Bottom left): meta classification rates for $\sIoU=0,>0$. The blue line are the LASSO fits for different $\lambda$ values, the orange line shows the performance of regular logistic regression fits ($\lambda=0$) where the input metrics are only those that have non-zero coefficients in the LASSO fit for the current $\lambda$. (Bottom right) same as bottom left, but for AUROC. The vertical dashed lines indicate the $\lambda$ value for which we obtained the best validation accuracy.} \label{fig:lasso1}
\end{figure*}

\begin{figure*}[t]
\centerline{\includegraphics[width=.79\textwidth]{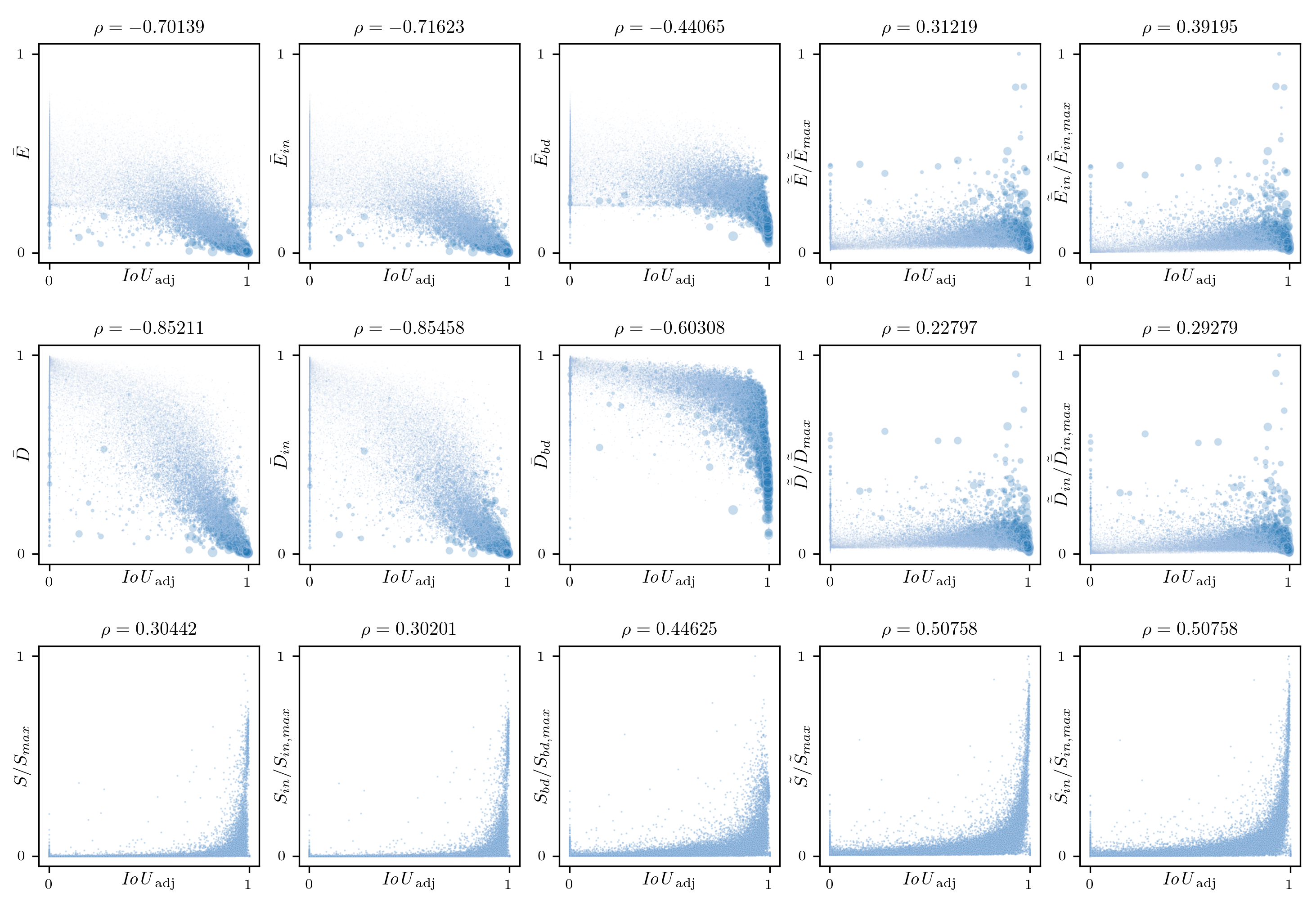}}
\caption{Correlations between $\sIoU$ and rescaled features for the DeepLabv3+Xception65 network. Dot sizes in the first two columns are proportional to $S$.} \label{fig:correlation1}
\end{figure*}

We investigate the properties of the metrics defined in the previous section \commentPS{for the example of a semantic segmentation of street scenes. To this end,} we consider the DeepLabv3+ network \cite{deeplab} for which \commentPS{we use} a reference implementation in Tensorflow \cite{tensorflow2015-whitepaper} as well as weights pretrained on the Cityscapes dataset \cite{cityscapes} \commentPS{and available on GitHub}.
The DeepLabv3+ implementation and weights are available for two network backbones: Xception65, which is a modified version of Xception \cite{xception} and is a powerful structure intended for server-side deployment, and MobilenetV2 \cite{mobilenet}, a fast structure designed for mobile devices. Each of these implementations have parameters tuning the segmentation accuracy. We choose the following best (for Xception65) and worst (for MobilenetV2) parameters in order to perform our analysis on two very distinct networks. Note, that the parameter set for the Xception65 setting also includes the evaluation of the input on multiple scales (averaging the results) which increases the accuracy and also leverages classification uncertainty. We refer to \cite{deeplab} for a detailed explanation of the chosen parameters.
\begin{itemize}
   \setlength\itemsep{-0.05em}
    \item DeepLabv3+Xception65: output stride $8$, decoder output stride $4$, evaluation on input scales $0.75, 1.00, 1.25$ -- $\mIoU = 79.72\%$ on the Cityscapes validation set
    \item DeepLabv3+MobilenetV2: output stride $16$, evaluation on input scale $1.00$ -- $\mIoU = 61.85\%$ on the Cityscapes validation set
\end{itemize}
Example segmentations and heat maps of the two networks are displayed in \cref{fig:heat}. For both networks, we consider the output probabilities and predictions on the Cityscapes validation set, which consists of 500 street scene images at a resolution of $2048\times 1024$. We compute the 15 constructed metrics as well as $\sIoU$ for each segment in the segmentations of the images. In order to investigate the predictive power of the metrics, we first compute the Pearson correlation $\rho \in [-1,1]$ between each feature and $\sIoU$. We report the results of this analysis in \cref{tab:correlations} and provide scatter plots of all features relative to $\sIoU$ in \cref{fig:correlation1}. Note, that in all computations, we only consider connected components with non-empty interior.

For both networks $\sIoU$ shows strong correlation with the mean distances $\bar D$ and $\bar D_\intr$ as well as with the mean entropies $\bar E$ and $\bar E_\intr$. On the other hand, the relative counterparts are less correlated with $\sIoU$. The relative segment size $\tilde S$ for the DeepLabv3+MobilenetV2 network shows a clear correlation whereas this is not the case for the more powerful DeepLabv3+Xception65 network.

\commentPS{In order to find more indicative measures, we now} investigate the predictive power of the metrics when they are combined.
% \comment{TODO: Naive Baseline Explanation}
For the Xception65 net, we obtain $45,\!194$ segments with non-empty interior of which $11,\!331$ have $\sIoU=0$. For the weaker MobilenetV2 this ratio is $42,\!261/17,\!671$. We would first like to detect segments with $\sIoU=0$, i.e., learn the meta classification \commentPS{task of identifying false positive segments} based on our 15 metrics and the segment-wise averaged \commentTH{probability distribution} vectors. We term these (standardized) inputs \commentPS{$x_k$ for a segment $k$. Further, let $y_k = \mathit{ceil}(\sIoU) = \{ 0 \text{ if } \sIoU=0, 1 \text{ if } \sIoU>0 \}$.%  \begin{align}
%    y_k=\left\{\begin{array}{ll}0&\text{if}~\sIoU=0\\1&\text{if}~\sIoU>0\end{array}\right.
%\end{align}
}
The least absolute shrinkage and selection operator (LASSO, \cite{Tibs1996}) is a popular tool for investigating the predictive power of different combinations of input variables. We compute a series of LASSO fits, i.e., $\ell_1$-penalized logistic regression fits 
\begin{align}
\begin{split}
    %  \min_{w} \left[ \sum_i - y_i \log ( \tau( w^T x_i ) ) - (1-y_i)(1-\log ( \tau( w^T x_i ) )) + \lambda \| w \|_1 \right]\ ,
    \min_{w} \sum_i & [- y_i \log ( \tau( w^T x_i ) ) \\ 
    & - (1-y_i)(1-\log ( \tau( w^T x_i ) )) + \lambda \| w \|_1 ]\ ,
\end{split}
\end{align}
for different regularization parameters $\lambda$ and standardized inputs (zero mean and unit standard deviation). Here, $\tau(\cdot)$ is the logistic function. Results for the Xception65 net are shown in~\cref{fig:lasso1}.

The top left and top right panels show, in which order the weight coefficients $w$ for each metric/predicted class become active. At the same time the bottom left and bottom right panels show, which weight coefficient causes which amount of increase in predictive performance in terms of meta-classification rate and AUROC, respectively. The AUROC is obtained by varying the decision threshold of the logistic regression output for deciding whether $\IoU=0$ or $\IoU>0$.

The first non-zero coefficient activates the ${\bar D}_\intr$ metric, which elevates the predictive power above our reference benchmark of choice, the mean entropy per component ${\bar E}$.
% ,  \commentPS{which we term \textit{entropy baseline}}.
Another significant gain is achieved when ${\bar D}_{\bdr}$ and the predicted classes come into play.
We obtain a meta-classification validation accuracy of up to $81.91\%(\pm0.13\%)$ and an AUROC of up to $87.71\%(\pm0.15\%)$ for Xception65. And also for the weaker MobilenetV2 we obtain $78.93\%(\pm0.17\%)$ classification accuracy and $86.77\%(\pm0.17\%)$ AUROC. 
% \comemnt{Comparing to the naive baselines }
We randomly choose 10 50/50 training/validation data splits and average the results, the numbers in brackets denote standard deviations of the averages.
%
%\comment{Note, that the chosen reference benchmark differs from the naive approach for the meta-classification task. The naive baseline would be the performance when all segments are assigned with maximal confidence to the majority class, i.e., the class $\sIoU=0$ or $\sIoU>0$ that has more samples. Let $\hat{\mathcal K } := \bigcup_x\hat{\mathcal K}_x$ be the set of all predicted segments in the whole dataset. The naive baseline for the accuracy can then be calculated via 
%$
%    \max\{ |\hat{\mathcal K}| - \sum_{k \in \hat{\mathcal K}} y_k, \sum_{k \in \hat{\mathcal K}} y_k \} / |\hat{\mathcal K}|
%$
%and the corresponding AUROC lies at $50\%$.}
% Another significant gain is achieved when ${\bar D}_{\bdr}$ and the predicted classes come into play. Noteworthily 
%We, \comment{however}, obtain a meta-classification validation accuracy of up to $81.91\%(\pm0.13\%)$ and an AUROC of up to $87.71\%(\pm0.15\%)$ for Xception65. And also for the weaker MobilenetV2 we obtain $78.93\%(\pm0.17\%)$ classification accuracy and $86.77\%(\pm0.17\%)$ AUROC. 
% \comemnt{Comparing to the naive baselines }
%We randomly choose 10 50/50 training/validation data splits and average the results, the numbers in brackets denote standard deviations of the averages.
%
%
Additionally, the bottom line of \cref{fig:lasso1} shows that there is almost no performance loss when only incorporating some of the metrics proposed by the LASSO trajectory. For both networks the classification accuracy corresponds to a logistic regression trained with unbalanced meta-classes $\sIoU=0$ and $\sIoU>0$, i.e., we did not adjust the class weights. On average (over the 10 training/validation splits) $6851$ \commentPS{components with vanishing $\sIoU$} are detected for Xception65 while $4480$ remain undetected, for MobilenetV2 this ratio is $14976$/$2695$. These ratios can be adjusted by varying the probability thresholds for deciding between $\sIoU=0$ and $\sIoU>0$. For this reason we state results in terms of AUROC \commentPS{which is threshold independent}.
\comment{We compare our results with two different baselines in \cref{tab:summary1}. The naive baseline is given by random guessing (randomly assigning a probability to each segment $k$ and then thresholding on it). The best meta classification accuracy is achieved for the threshold being either $0$ or $1$. For $ I_0 = | \{ k : \sIoU=0\} |$ and  $I_1 = | \{ k : \sIoU>0\}|$ the naive baseline accuracy is then given by $\frac{\mathrm{max}(I_0,I_1)}{I_0+I_1}$. The corresponding AUROC value is $50\%$. Another baseline is to equip our approach only with a single  metric. For this purpose we choose the entropy as it is commonly used for uncertainty quantification.}

\begin{figure*}[t]
\begin{center}
\begin{minipage}{.5\textwidth}
  \centering
  \includegraphics[width=.99\linewidth]{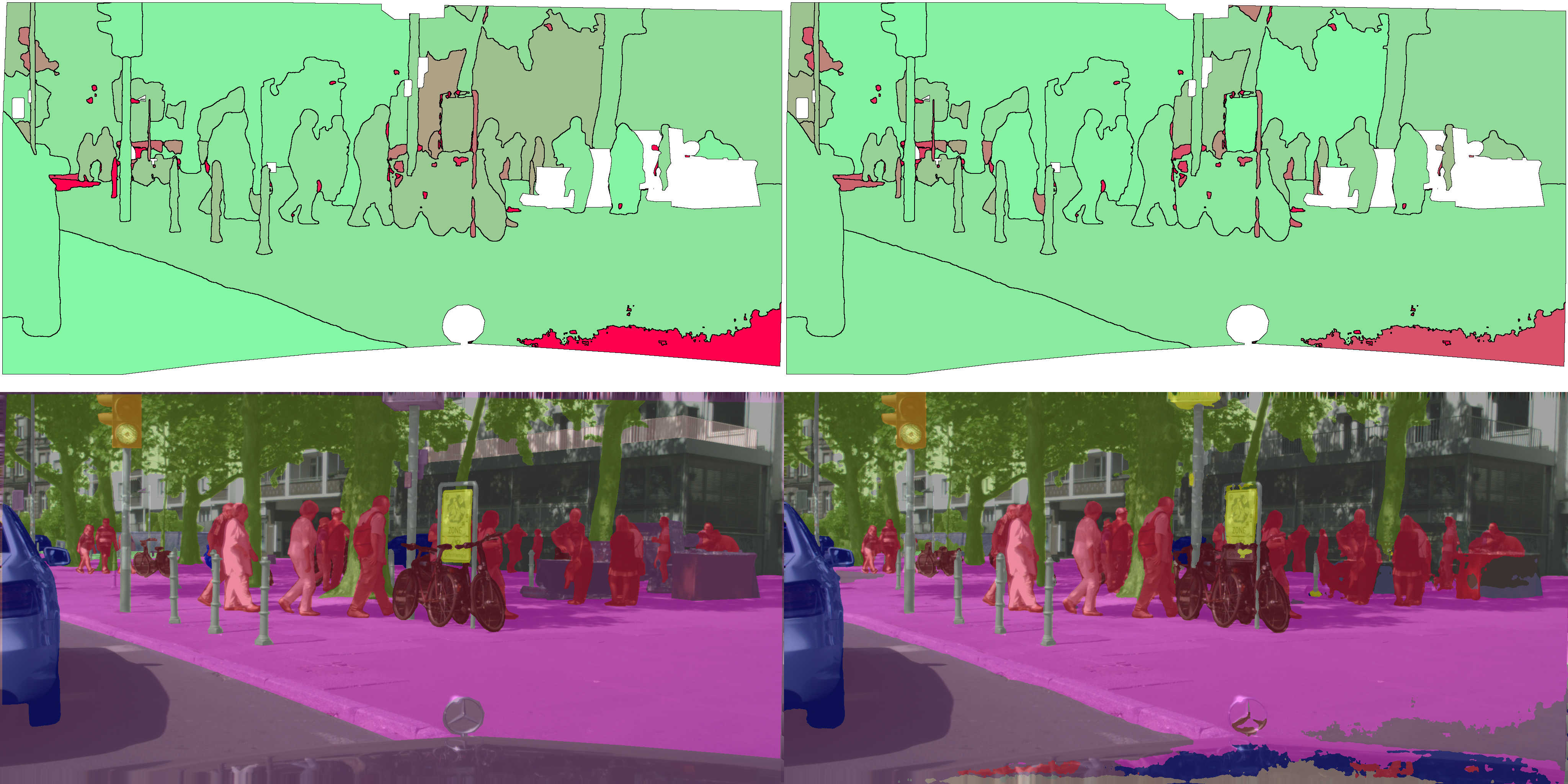}
  \small (a) DeepLabv3+ Xception65
\end{minipage}%
\begin{minipage}{.5\textwidth}
  \centering
  \includegraphics[width=.99\linewidth]{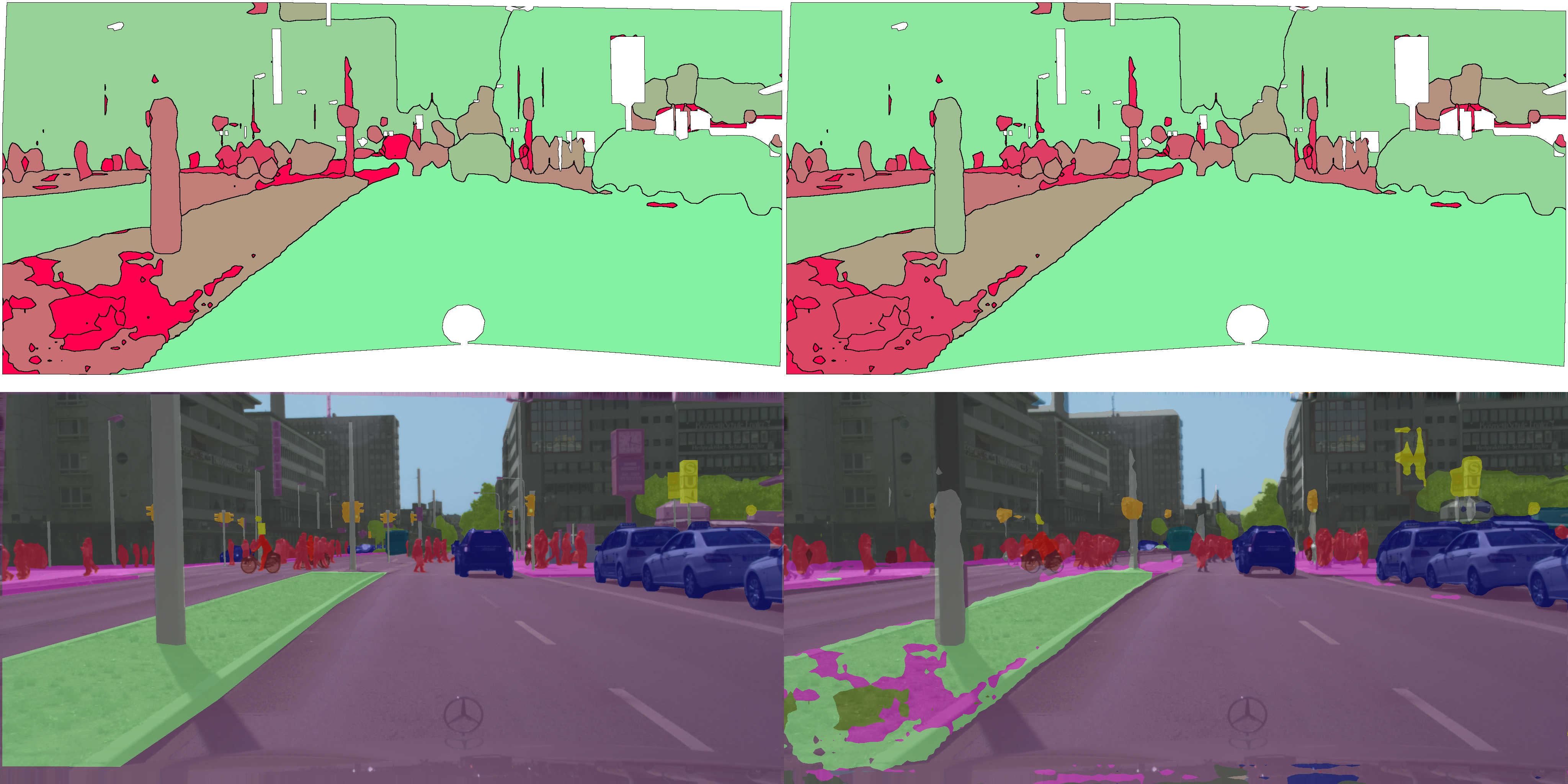}
  \small (b) DeepLabv3+ MobilenetV2
\end{minipage}%
\end{center}
\caption{Prediction of the $\sIoU$ with linear regression. Each of the two sub-figures (a) and (b) consists of ground truth (bottom left), predicted segments (bottom right), true $\sIoU$ for the predicted segments (top left) and predicted $\sIoU$ for the predicted segments (top right). In the top row, green color corresponds to high $\sIoU$ values and red color to low ones, for the white regions there is no ground truth available. These regions are excluded from the statistical evaluation.} \label{fig:illus1}
\end{figure*}

\begin{table*}[htb]
\begin{center}
\scalebox{0.77}{
\begin{tabular}{|l||r|r||r|r|}
\hline
& \multicolumn{2}{|c||}{Xception65} & \multicolumn{2}{|c|}{MobilenetV2}  \\
\hline
Cityscapes & \multicolumn{1}{|c|}{training} & \multicolumn{1}{|c||}{validation} & \multicolumn{1}{|c|}{training} & \multicolumn{1}{|c|}{validation} \\
\hline
& \multicolumn{4}{|c|}{Classification $\sIoU=0,>0$} \\
\hline
ACC, penalized             & $81.88\%(\pm0.13\%)$ & $81.91\%(\pm0.13\%)$ & $78.87\%(\pm0.13\%)$ & $78.93\%(\pm0.17\%)$ \\
ACC, unpenalized           & $81.91\%(\pm0.12\%)$ & $81.92\%(\pm0.12\%)$ & $78.84\%(\pm0.14\%)$ & $78.93\%(\pm0.18\%)$ \\
ACC, entropy only      & $76.36\%(\pm0.17\%)$ & $76.32\%(\pm0.17\%)$ & $68.33\%(\pm0.27\%)$ & $68.57\%(\pm0.25\%)$ \\
\hline
ACC, naive baseline        & \multicolumn{2}{|c||}{$74.93\%$} & \multicolumn{2}{|c|}{$58.19\%$} \\
\hline
AUROC, penalized           & $87.71\%(\pm0.14\%)$ & $87.71\%(\pm0.15\%)$ & $86.74\%(\pm0.18\%)$ & $86.77\%(\pm0.17\%)$ \\
AUROC, unpenalized         & $87.72\%(\pm0.14\%)$ & $87.72\%(\pm0.15\%)$ & $86.74\%(\pm0.18\%)$ & $86.76\%(\pm0.18\%)$ \\
AUROC, entropy only    & $77.81\%(\pm0.16\%)$ & $77.94\%(\pm0.15\%)$ & $76.63\%(\pm0.24\%)$ & $76.74\%(\pm0.24\%)$ \\
\hline
% AUROC, naive baseline        & \multicolumn{2}{|c||}{$50.00\%$} & \multicolumn{2}{|c|}{$50.00\%$} \\
% \hline
& \multicolumn{4}{|c|}{Regression $\sIoU$} \\
\hline
$\sigma$, all metrics      & $0.181(\pm0.001)$    & $0.182(\pm0.001)$    & $0.130(\pm0.001)$    & $0.130(\pm0.001)$    \\ 
$\sigma$, entropy only     & $0.258(\pm0.001)$    & $0.259(\pm0.001)$    & $0.215(\pm0.001)$    & $0.215(\pm0.001)$    \\ 
$R^2$, all metrics         & $75.06\%(\pm0.22\%)$ & $74.97\%(\pm0.22\%)$ & $81.50\%(\pm0.23\%)$ & $81.48\%(\pm0.23\%)$ \\ 
$R^2$, entropy only        & $49.37\%(\pm0.32\%)$ & $49.02\%(\pm0.32\%)$ & $49.32\%(\pm0.31\%)$ & $49.12\%(\pm0.32\%)$ \\ 
\hline
\end{tabular}
}
\end{center}
\caption{Summarized results for classification and regression \commentRC{for Cityscapes}, averaged over 10 runs. The numbers in brackets denote standard deviations of the computed mean values.} \label{tab:summary1}
\end{table*}

Ultimately, we want to predict $\sIoU$ values for all connected components and thus model an uncertainty measure. We now resign from regularization and use a linear regression model to predict the $\sIoU$. \Cref{fig:regr} depicts the quality of a single linear regression fit for each of the two segmentation networks. For MobilenetV2 we obtain an $R^2$ value of $81.48\%(\pm0.23\%)$ and for Xception65 $74.93\%(\pm0.22\%)$. \Cref{fig:illus1} illustrates the constructed uncertainty measure with two showcases. Averaged results over 10 runs including standard deviations $\sigma$ and previous meta classification result are summarized in \cref{tab:summary1}. In all cases, the presented approach clearly outperforms the entropy.
%\comment{only benchmark.}.
The linear regression models do not overfit the data and note-worthily we obtain prediction standard deviations of down to $0.130$ and almost no standard deviation for the averages. The classification accuracy and AUROC results are slightly biased towards the validation results as they correspond to the particular $\lambda$ value that maximizes the validation accuracy. 
% An additional discussion on the difference (also in performance between) $\sIoU$ and $\IoU$ can be found in the appendix.

% \begin{figure*}[t]
% (a) DeepLabv3+ MobilenetV2 \\
% \centerline{\includegraphics[width=1.0\textwidth]{./figs/MN_img53.jpg}}
% \vspace{1ex}
% (b) DeepLabv3+ Xception65 \\
% \centerline{\includegraphics[width=1.0\textwidth]{./figs/XC_img100.jpg}}
% \caption{Prediction of the $\sIoU$ with linear regression. Each of the two sub-figures (a) and (b) consists of ground truth (bottom left), predicted segments (bottom right), true $\sIoU$ for the predicted segments (top left) and predicted $\sIoU$ for the predicted segments (top right). In the top row, green color corresponds to high $\sIoU$ values and red color to low ones, for the white regions there is no ground truth available. These regions are excluded from the statistical evaluation.} \label{fig:illus1}
% \end{figure*}

\section{Numerical Experiments: Brain Tumor Segmentation}

\begin{table*}[htb]
\begin{center}
\scalebox{0.77}{
% \begin{tabular}{|l||c|c|c||c|c|c|c||c|}
% \hline
% Metric & \multicolumn{3}{|c||}{Dice Coefficient} & \multicolumn{5}{|c|}{Intersection over Union} \\
% \hline
% Network   & WT & TC & ET & BG & NT & PE & ET & $\mIoU$  \\
% \hline
% 2D U-Net by Kermi et al. & $88.09\%$ & $77.38\%$ & $78.89\%$ & $99.74\%$ & $46.30\%$ & $58.16\%$ & $65.14\%$ & $67.14\%$  \\
% \hline
% 3D Net by Myronenko et al. & $88.83\%$ & $81.07\%$ & $79.63\%$ & $99.74\%$ & $53.53\%$ & $59.14\%$ & $66.15\%$ & $69.64\%$  \\
% \hline
% \end{tabular}
\begin{tabular}{|l||c|c|c||c||c|c|c||c|}
\hline
Metric & \multicolumn{4}{|c||}{Dice Coefficient} & \multicolumn{4}{|c|}{Intersection over Union} \\
\hline
& WT & TC & ET & $\mathit{mDice}$ & \multicolumn{2}{|c||}{$\mIoU$ nested} & \multicolumn{2}{|c|}{$\mIoU$ single} \\
\hline
2D U-Net by Kermi et al. & $88.09\%$ & $77.38\%$ & $78.89\%$ & $81.45\%$ & \multicolumn{2}{|c||}{$68.99\%$} & \multicolumn{2}{|c|}{$67.14\%$}  \\
\hline
3D Net by Myronenko et al. & $88.83\%$ & $81.07\%$ & $79.63\%$ & $83.18\%$ & \multicolumn{2}{|c||}{$71.40\%$} & \multicolumn{2}{|c|}{$69.64\%$}   \\
\hline
\end{tabular}
}
\end{center}
\caption{\commentRC{Evaluation scores on validation split. The nested classes whole tumor (WT), tumor core (TC) and enhancing tumor (ET) are evaluated with the Dice coefficient. For comparison purposes, the mean Dice score is reported as well as mean Intersection over Union for nested classes and single classes (background, non-enhancing tumor, peritumoral edema and enhancing tumor).}} \label{tab:benchmark_brats1}
\end{table*}

\begin{table*}[htb]
\begin{center}
\scalebox{0.77}{
\begin{tabular}{|l||r|r||r|r|}
\hline
& \multicolumn{2}{|c||}{2D U-Net by Kermi et al.} & \multicolumn{2}{|c|}{3D Net by Myronenko et al.}  \\
\hline
BraTS2017 & \multicolumn{1}{|c|}{training} & \multicolumn{1}{|c||}{validation} & \multicolumn{1}{|c|}{training} & \multicolumn{1}{|c|}{validation} \\
\hline
& \multicolumn{4}{|c|}{Classification $\sIoU=0,>0$} \\
 \hline
ACC, penalized             & $89.30\%(\pm0.18\%)$ & $89.39\%(\pm0.17\%)$ & $88.40\%(\pm0.27\%)$ & $88.42\%(\pm0.27\%)$ \\
ACC, unpenalized           & $89.29\%(\pm0.19\%)$ & $89.40\%(\pm0.18\%)$ & $88.38\%(\pm0.27\%)$ & $88.40\%(\pm0.28\%)$ \\
ACC, entropy only      & $87.96\%(\pm0.12\%)$ & $87.96\%(\pm0.12\%)$ & $86.69\%(\pm0.20\%)$ & $86.69\%(\pm0.20\%)$ \\
\hline
ACC, naive baseline        & \multicolumn{2}{|c||}{$88.30\%$} & \multicolumn{2}{|c|}{$86.35\%$} \\
\hline
AUROC, penalized           & $91.84\%(\pm0.25\%)$ & $91.93\%(\pm0.24\%)$ & $91.51\%(\pm0.22\%)$ & $91.55\%(\pm0.22\%)$ \\
AUROC, unpenalized         & $91.83\%(\pm0.25\%)$ & $91.93\%(\pm0.24\%)$ & $91.49\%(\pm0.22\%)$ & $91.53\%(\pm0.22\%)$ \\
AUROC, entropy only    & $86.68\%(\pm0.25\%)$ & $86.73\%(\pm0.25\%)$ & $86.59\%(\pm0.28\%)$ & $86.74\%(\pm0.28\%)$ \\
\hline
& \multicolumn{4}{|c|}{Regression $\sIoU$} \\
\hline
$\sigma$, all metrics      & $0.148(\pm0.001)$    & $0.149(\pm0.001)$    & $0.171(\pm0.001)$    & $0.171(\pm0.001)$ \\ 
$\sigma$, entropy only & $0.178(\pm0.001)$    & $0.178(\pm0.001)$    & $0.198(\pm0.001)$    & $0.197(\pm0.001)$ \\ 
$R^2$, all metrics         & $84.22\%(\pm0.21\%)$ & $84.15\%(\pm0.21\%)$ & $79.53\%(\pm0.28\%)$ & $79.64\%(\pm0.28\%)$ \\ 
$R^2$, entropy only    & $77.18\%(\pm0.18\%)$ & $77.30\%(\pm0.17\%)$ & $72.63\%(\pm0.27\%)$ & $72.91\%(\pm0.27\%)$ \\ 
\hline
\end{tabular}
}
\end{center}
\caption{Summarized results for classification and regression for BraTS2017, averaged over 10 runs. The numbers in brackets denote standard deviations of the computed mean values.} \label{tab:summary_brats1}
\end{table*}

\begin{figure}[tb]
\begin{center}
\begin{minipage}{.495\linewidth}
    \centering
    \includegraphics[width=.99\linewidth]{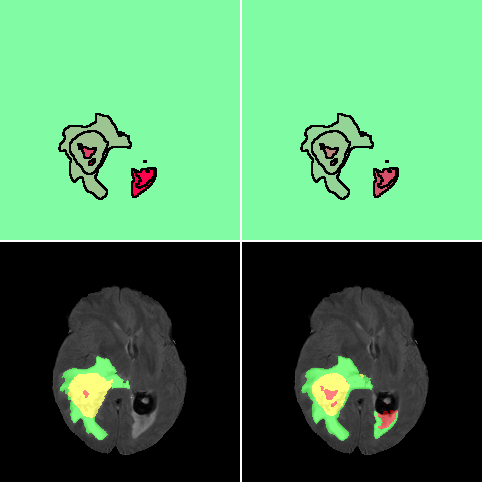}\\
    \small (a) 2D U-Net by Kermi et al.
\end{minipage}
\hfill
\begin{minipage}{.495\linewidth}
    \centering
    \includegraphics[width=.99\linewidth]{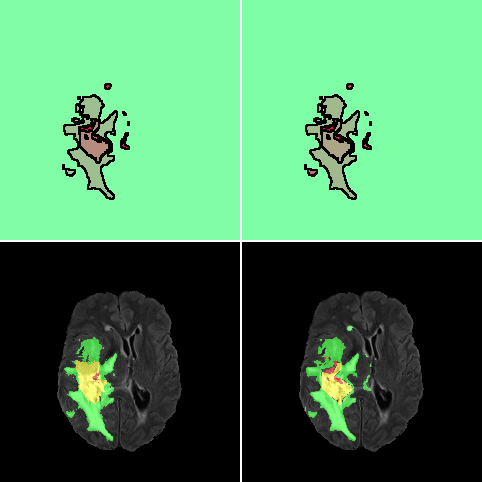}\\
    \small (b) 3D Net by Myronenko et al.
\end{minipage}
\end{center}
\caption{\commentRC{Two demonstrations (left and right four panels,  analogously to~\cref{fig:illus1}) of our method's performance of predicting $\sIoU$ on BraTS2017.
The whole tumor (WT) includes all colored segments (union of green, yellow \& red), the tumor core (TC) the yellow joined with the red colored segments and the enhancing tumor (ET) only the yellow colored segments.}}
\label{fig:brats1}
\end{figure}

\changed{The method we propose only uses dispersion heat maps and softmax probabilities as inputs. Any additional heat map increases the performance as long as there is no overfitting. 
%Thus, the generalization of our approach across different datasets is to be expected. 
\commentRC{Thus, we expect our approach to generalize across different datasets even from different domains.}
To demonstrate this, we perform additional tests with the brain tumor segmentation dataset BraTS2017~\cite{Bakas2017,6975210} and two different networks,}
\commentRC{i.e., a simple 2D network and a more complex 3D network.}
% \commentRC{, i.e. 1XXXX~\cite{kermi2018deep} and 2XXXX~\cite{Myronenko18}.}
% \commentRC{that are both based on the U-Net~\cite{10.1007/978-3-319-24574-4_28}, originally well-known for its performance on biomedical image segmentation.}
%, i.e., the U-Net~\cite{kermi2018deep} and XXXX~\cite{}.
\commentRC{Compared to \changed{the segmentation of street scenes, brain tumor segmentation involves way fewer classes. The background class is usually dominat}. In BraTS2017, around $98\%$ of all pixels \changed{are} background, the remaining classes comprise necrotic/non-enhancing tumor, peritumoral edema and enhancing tumor. \changed{For benchmarks of predictive methods, these labels} are combined into three nested classes: whole tumor (WT), tumor core (TC) and enhancing tumor (ET) (see~\cref{fig:brats1}). The most commonly used evaluation metric is the so-called \textit{Dice-Coefficient}~\cite{Zou2004} that is defined as $\dsc := 2\mathit{TP} / (2\mathit{TP}+\mathit{FP}+\mathit{FN})$ }
\changed{where $\mathit{TP}$, $\mathit{FP}$ and $\mathit{FN}$ denote all true positive, false positive and false negative pixels, respectively, for a chosen class.}
%$$
%\dsc(k) := \frac{2|k \cap K'|}{|k \cup K'|+|k \cap K'|}\,, \qquad K' = \bigcup_{k' \in {\mathcal K}_x|_k} k' .
%$$}
%instead of whole 3D scans.

\commentRC{The BraTS data is available as magnetic resonance imaging (MRI) brain scans from three viewing angles and with four modalities of higher grade gliomas (HGG) and lower grade gliomas (LGG).
For training and validation, we combine HGG and LGG images and randomly split the data 80/20. We train the networks from scratch with the different scan modalities stacked as network's input channels. 
%present numbers for the same tests as in the previous section, 
Once this is done, we perform tests analogously to the previous section. The performance of the two networks being used on our validation split are reported in \cref{tab:benchmark_brats1}, the results for classification and regression are summarized in \cref{tab:summary_brats1}.}

\commentRC{For the first test we use the network by Kermi et al.~\cite{kermi2018deep}. It is based on the U-Net~\cite{10.1007/978-3-319-24574-4_28} which is originally well-known for its performance on biomedical image segmentation. We train the network on randomly sampled 2D patches from axial (top view) slices of the brain scans.
%With this network we achieve dice scores of \textbf{88.09\%}, \textbf{77.38\%} and \textbf{78.89\%} for WT, TC and ET, respectively, on our validation split.
The results of our prediction rating methods are computed for 22,242 non-empty segments of which 2,603 have $\sIoU=0$.} 
\changed{Indeed, we obtain \commentRC{higher} accuracy values compared to
%the results for 
Cityscapes, however the gain over the single metric baseline is not as big. This is \commentRC{primarily} due to a strong} \commentRC{correlation between $E$ and $\sIoU$ ($-0.87794$). 
%At this point, we want to emphasize that considering only the entropy is already part of our approach, just in the simplest form. Comparing to the guessing accuracy baseline, which lies at 76.59\%, is more appropriate here.
\comment{In this case, the gain over the naive baseline is marginal. This may be misleading to the disadvantage of our method as the high naive accuracy is caused by the strong sample imbalance of the meta-classes. The corresponding AUROC value of $91.93\%$ shows that our method
%, as a matter of fact, 
meta-classifies with significant\commentMR{ly} higher confidence when incorporating all %dispersion
metrics.}
Regarding the $R^2$ value of our regression model for predicting $\sIoU$, we observe a gain from $77.30\%(\pm0.17\%)$ to $84.15\%(\pm0.21\%)$ when incorporating all metrics instead of only the entropy.}

\commentRC{Next, we compare the U-Net's performance to the state-of-the-art network by Myronenko et al.~\cite{Myronenko18}. One main difference is that the latter network considers the MRI scans' 3D contextual information by processing multiple contiguous 2D slices at once, i.e., we train the network on randomly sampled 3D patches. As a consequence, the model is more complex and the number of trainable parameters is noticeably increased (10.1M vs. 17.3M).
%but thereby we expect the network to have an improved performance on the task of brain tumor segmentation. 
%The dice scores for this network are \textbf{88.83\%}, \textbf{81.07\%} and \textbf{79.63\%} for WT, TC and ET, respectively. 
We perform the evaluation in the same \changed{2D} slice-wise manner as for the U-Net. The results are now computed for 24,397 non-empty segments of which 3331 have $\sIoU=0$. 
%The guessing accuracy baseline lies at 86.34\%. 
Again, we observe a strong correlation between $E$ and $\sIoU$ of $-0.84294$ which results in a nearly identical gain in terms of percent points over the single metric baseline as for the U-Net. Also with respect to the $R^2$ value of our regression model, the gain is again around 7\%, whereas the absolute value with $79.64\%(\pm0.28\%)$ for all metrics is not as high as for the U-Net.} 
\comment{All results are summarized in \cref{tab:summary_brats1}, the visualization in~\cref{fig:brats1} further demonstrates the performance of our method.}

\section{Adjusted Intersection over Union} \label{sec:sIou}

\begin{table*}[htb]
\begin{center}
\scalebox{0.80}{
\begin{tabular}{|l||r|r||r|r|}
\hline
& \multicolumn{2}{|c||}{Xception65} & \multicolumn{2}{|c|}{MobilenetV2}  \\
\hline
& \multicolumn{1}{|c|}{training} & \multicolumn{1}{|c||}{validation} & \multicolumn{1}{|c|}{training} & \multicolumn{1}{|c|}{validation} \\
\hline
& \multicolumn{4}{|c|}{Regression $\sIoU$} \\
\hline
$\sigma$, all metrics       & $0.181(\pm0.001)$    & $0.182(\pm0.001)$    & $0.130(\pm0.001)$    & $0.130(\pm0.001)$    \\
$\sigma$, entropy only      & $0.258(\pm0.001)$    & $0.259(\pm0.001)$    & $0.215(\pm0.001)$    & $0.215(\pm0.001)$    \\
$R^2$, all metrics          & $75.06\%(\pm0.22\%)$ & $74.97\%(\pm0.22\%)$ & $81.50\%(\pm0.23\%)$ & $81.48\%(\pm0.23\%)$ \\
$R^2$, entropy only         & $49.37\%(\pm0.32\%)$ & $49.02\%(\pm0.32\%)$ & $49.32\%(\pm0.31\%)$ & $49.12\%(\pm0.32\%)$ \\
\hline
& \multicolumn{4}{|c|}{Regression $\IoU$} \\
\hline
$\sigma$, all metrics       & $0.192(\pm0.001)$    & $0.192(\pm0.001)$    & $0.135(\pm0.001)$    & $0.135(\pm0.001)$    \\
$\sigma$, entropy only      & $0.267(\pm0.001)$    & $0.268(\pm0.001)$    & $0.217(\pm0.001)$    & $0.217(\pm0.001)$    \\
$R^2$, all metrics          & $72.90\%(\pm0.21\%)$ & $72.77\%(\pm0.21\%)$ & $79.63\%(\pm0.27\%)$ & $79.58\%(\pm0.27\%)$ \\
$R^2$, entropy only         & $47.43\%(\pm0.28\%)$ & $47.07\%(\pm0.28\%)$ & $47.73\%(\pm0.37\%)$ & $47.50\%(\pm0.38\%)$ \\
\hline
\end{tabular}
}
\end{center}
\caption{Comparison of regression results for segment-wise fitting $\sIoU$ and $\IoU$, averaged over 10 runs. The numbers in brackets denote standard deviations of the computed mean values.} \label{tab:summary2}
\end{table*}

% \begin{figure*}[htb]
% \centerline{\includegraphics[height=89pt]{./figs/sketch_pt1.png} \hfill
% \includegraphics[height=89pt]{./figs/sketch_pt2.png} \hfill
% \includegraphics[height=89pt]{./figs/sketch_pt3.png} \hfill
% \includegraphics[height=89pt]{./figs/sketch_pt5.png}}
% \caption{Illustration of the different behaviors of $\IoU$ and $\sIoU$. We have $\sIoU$ per segment (left panel), $\IoU$ per segment (center left), ground truth (center right) and detail views for the crucial area of the predicted segmentation (top right) and the corresponding ground truth (bottom right), green stands for high $\IoU$ and $\sIoU$ values, red for low ones, respectively. The top right panel shows that the prediction for the class `nature' is decoupled into two components by the traffic light's prediction. The $\IoU$ rates this small part on the left from the traffic light very badly even though the prediction is absolutely fine. The adjusted $\sIoU$ circumvents this type of problems.} \label{fig:sketch}
% \end{figure*}

\begin{figure*}[htb]
\begin{center}
%\hfill
\begin{minipage}{0.2\textwidth}
\centering
\includegraphics[height=77pt]{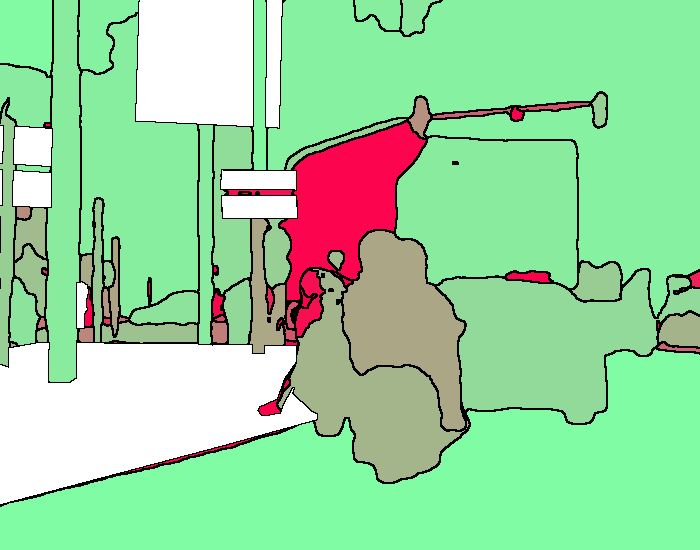}\\
\small (a)
\end{minipage}
\begin{minipage}{0.30\textwidth}
\centering
\includegraphics[height=77pt]{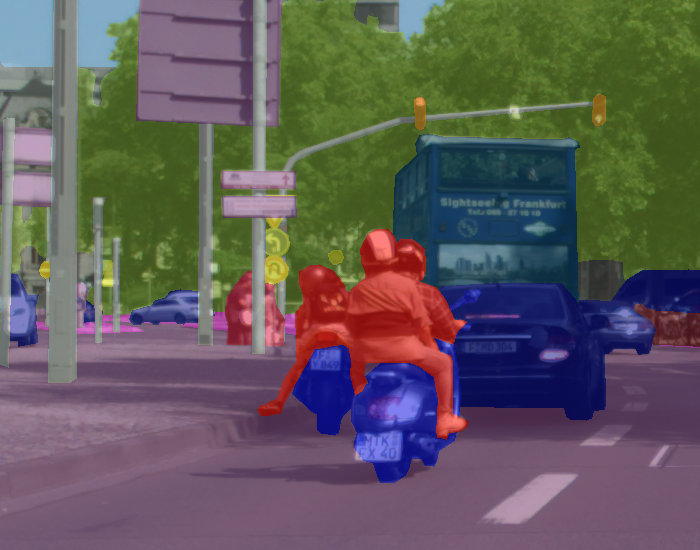}
\includegraphics[height=77pt]{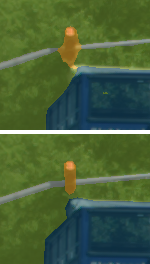}\\
\small (b)
\end{minipage}
\begin{minipage}{0.2\textwidth}
\centering
\includegraphics[height=77pt]{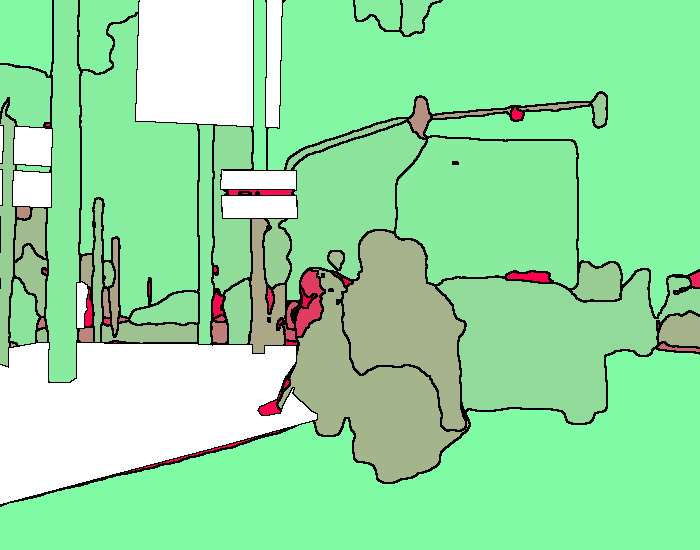}\\
\small (c)
\end{minipage}
%\begin{minipage}{.5\textwidth}
%  \centering
%  \includegraphics[width=.99\linewidth]{./figs/XC_img100.jpg}
%  \small (a) DeepLabv3+ Xception65
%\end{minipage}%
%\includegraphics[height=77pt]{./figs/sketch_pt2.png}\hspace{1ex}%
%\includegraphics[height=77pt]{./figs/sketch_pt3.png}\hspace{1ex}%
%\includegraphics[height=77pt]{./figs/sketch_pt5.png}\hspace{1ex}%
%\includegraphics[height=77pt]{./figs/sketch_pt1.png}%
%\hfill
\end{center}
\caption{Illustration of the different behaviors of $\IoU$ and $\sIoU$. We have (a): $\IoU$ per segment, (b) left: ground truth, right: detail views for the crucial area of the predicted segmentation (top) and the corresponding ground truth (bottom) and (c): $\sIoU$ per segment. Green stands for high $\IoU$ and $\sIoU$ values, red for low ones, respectively. The top right panel in (b) shows that the prediction for the class `nature' is decoupled into two components by the traffic light's prediction. The $\IoU$ rates this small part on the left from the traffic light very badly even though the prediction is absolutely fine. The adjusted $\sIoU$ circumvents this type of problems.} \label{fig:sketch}
\end{figure*}

In \cref{sec:pixel} we introduced the adjusted $\sIoU(k)$ for an inferred segment $k \in \hat{\mathcal{K}}_x$ which slightly deviates from the ordinary $\IoU(k)$. The reason for this is the following: In some cases it can happen that a connected component in the ground truth is split into two or more in the prediction. 
% Imagine a pole on the sidewalk that is labeled as sidewalk in the ground truth, but detected by the neural network and thus splits the sidewalk segment into two components. 
Each component would be assigned a moderate $\IoU$ value even though they are predicted very well (cf.~also~\cref{fig:sketch}). For this reason we introduce the adjusted $\sIoU$ that does not depreciate the prediction of a segment if the remainder of the ground truth is well covered by other predicted segments belonging to the same class.

Clearly, we have $\sIoU(k) = \IoU(k) = 1$ if and only if the predicted segment $k$ and the ground truth $K'$ match for each pixel, $\sIoU=\IoU=|k \cap K'|=0$ when ground truth and predicted segment do not overlap, i.e., $k \cap K' = \emptyset$, and it holds $\sIoU \geq \IoU$. Thus, the classification task is invariant under interchanging $\IoU$ and $\sIoU$, however, the regression task is not.

Carrying out the regression tests from \cref{sec:numexp} for the $\sIoU$ with the $\IoU$ as well, we observe that the regression fit for the $\sIoU$ achieves $R^2$ values that are roughly $2\%$ higher than those for the $\IoU$, cf.~\cref{tab:summary2}. Usually, for performance measures in semantic segmentation, the $\IoU$ is computed for a chosen class over the whole image. This means that each pixel of the union of prediction and ground truth is only counted once in the denominator of the $\IoU$. On the other hand, a ground truth pixel may contribute to $\IoU$s of several segments. In this sense, in the context of semantic segmentation, the adjusted $\sIoU$ is closer in spirit to the regular image-wise $\IoU$.

It seems natural to consider an intersection over segment size $\mathit{IoS}(k) = |k \cap K'| / S$ as well. However, $\mathit{IoS}(k) = 1$ for a segment $k$ does not imply that a segment perfectly matches the corresponding ground truth. Consequently, one should refrain from considering this measure, at least as an exclusively used performance measure.

\section{Conclusion and Outlook}

We have shown statistically that per-segment metrics derived from entropy, probability difference, segment size and the predicted class clearly contain information about the reliability of the segments and constructed an approach for detecting unreliable segments in the network's prediction. In our tests with publicly available networks and datasets, Cityscapes and BraTS2017, the computed logistic LASSO fits for meta classification task $\sIoU=0$ vs.\ $\sIoU>0$ \changed{achieve AUROC values of up to $91.55\%$}. When predicting the $\sIoU$ with a linear regression fit we obtain a prediction standard deviation of down to $0.130$, as well as $R^2$ values of up to \changed{$84.15\%$}.
These results could be further improved when incorporating model uncertainty in heat map generation. We believe that using MC dropout will further improve these results, just like the development of ever more accurate networks. We plan to use our method for detecting labeling errors, for label acquisition in active learning and we plan to investigate further metrics that may leverage detection accuracy. Apart from that, detection mechanisms built on the softmax input and even earlier layers could be thought of. The source code of our method is publicly available at \ifwacvfinal\href{https://github.com/mrottmann/MetaSeg}{\texttt{https://github.com/mrottmann/MetaSeg}}\else{}\texttt{TBA}\fi.

\paragraph{Acknowledgements.}
This work is funded in part by \ifwacvfinal Volkswagen Group Innovation\else{}TBA\fi.

{\small
\bibliographystyle{ieee}
\bibliography{egbib}
}

\end{document}